# A Transformer-Based Multi-Stream Approach for Isolated Iranian Sign Language Recognition

Ali Ghadami[1,2], Alireza Taheri[1,*], Ali Meghdari[1]

[1] *Department of Mechanical Engineering, Sharif University of Technology, Tehran, Iran*
[2] *Present Address: Department of Mechanical Engineering, University of Michigan, Ann Arbor, USA*

**Abstract**

Sign language is an essential means of communication for millions of people around the world and serves as their primary language. However, most communication tools are developed for spoken and written languages which can cause problems and difficulties for the deaf and hard of hearing community. By developing a sign language recognition system, we can bridge this communication gap and enable people who use sign language as their main form of expression to better communicate with people and their surroundings. This recognition system increases the quality of health services, improves public services, and creates equal opportunities for the deaf community. This research aims to recognize Iranian Sign Language words with the help of the latest deep learning tools such as transformers. The dataset used includes 101 Iranian Sign Language words frequently used in academic environments such as universities. The network used is a combination of early fusion and late fusion transformer encoder-based networks optimized with the help of genetic algorithm. The selected features to train this network include hands and lips key points, and the distance and angle between hands extracted from the sign videos. Also, in addition to the training model for the classes, the embedding vectors of words are used as multi-task learning to have smoother and more efficient training. This model was also tested on sentences generated from our word dataset using a windowing technique for sentence translation. Finally, the sign language training software that provides real-time feedback to users with the help of the developed model, which has 90.2% accuracy on test data, was introduced, and in a survey, the effectiveness and efficiency of this type of sign language learning software and the impact of feedback were investigated. This software, and this research in general, can be an initial step in the practical implementation of sign language recognition models in the real world, which can greatly help the deaf community.

## 1. Introduction

According to the World Health Organization (WHO), about 466 million deaf and hard-of-hearing people live in the world. Although sign language is not the main means of communication for all of them, among this group, more than 72 million deaf people use 300 different types of sign language, and as a result, sign language is the main language and method of communication among millions of people on the planet [1]. There are a number of reasons why sign language translation systems are important. For example, translating sign language television content, facilitating communication between deaf and hearing people, and developing sign language interpreter robots to interact with deaf people. Achieving such a system with high accuracy is a challenging problem and highlights the importance of continuous development of tools and methods to solve this problem.

Sign language translation is generally divided into two categories: continuous recognition and isolated recognition. In the context of isolated sign language translation, the model receives input in the form of a video or information featuring a single gesture, such as a single sign language word, and the objective is to translate individual gestures. Unlike isolated translation, where the goal is to translate single words, in continuous translation, the aim is to translate a sentence that includes any number of words. As it seems, continuous translation is more complicated than isolated translation because the boundaries of the words in the video or input signal must be determined and then the translation of single words should be done. Of course, this is not the only method, and researchers have tried to translate the entire text without intermediaries, which have also obtained favorable results [2], [3].

From the perspective of the data used for sign language recognition, the existing methods are categorized into two categories, which are methods based on sensors connected to the person, such as gloves, and methods based on vision. It should be noted that due to the limitations of the sensors attached to the person, researchers in this field have moved towards vision-based approaches. Among these limitations, we can mention the cost of these gloves, the need for additional cumbersome equipment, and the inability to capture all the necessary features.

Many approaches have been investigated to solve the problem of sign language recognition, among which we can mention older classical methods, such as statistical methods and machine learning, as well as recently developed deep learning methods. In [4], researchers introduced the Support Vector Machine (SVM) as a suitable

*Corresponding author: artaheri@sharif.edu , Tel: +982166165531*

and efficient algorithm for real-time sign language classification. Additionally, Chong and Lee developed an American Sign Language (ASL) recognition system using SVM and deep learning. In this research, for the 26 letters of sign language, using this algorithm achieved the accuracy of 80.3%, and using deep learning, an accuracy of 93.81% was reached [5]. Additionally, in recent years, some researchers have tried to use the Hidden Markov Model (HMM) along with other methods to achieve better results. For example, by combining the Hidden Markov Model and Bi-LSTM, an accuracy of 97.85% for one-handed signs and 94.55% for two-handed signs was obtained [6].

With the increasing progress in deep learning, its application in sign language recognition has also become prominent, so most of the current efforts in this field are based on deep learning tools. The techniques used include Deep Belief Networks (DBN), Convolutional Neural Networks (CNN), Recurrent Neural Networks (RNN), Recurrent Convolutional Neural Networks (RCNN), and transformer networks.

Using 3D convolutional neural networks, Sharma and Kumar have succeeded in recognizing isolated American Sign Language (ASL) words with 96% accuracy for 100 words [7]. Daroya et al. used a convolutional neural network to classify RGB images of static hand gestures (representing a letter) related to sign language [8]. Fang et al. used a bi-directional RNN and LSTM for translation at the word and sentence level of sign language. The experimental result showed that the RNN model can successfully capture the important features of American Sign Language words [9]. In another research, Correia et al. introduced a novel Spatial-Temporal Graph Convolutional Network for sign language recognition, leveraging human skeletal movements to capture both spatial and temporal dynamics, while also providing a new dataset of human skeletons based on ASLLVD for further research in the field [10]. Ye et al. introduced a hybrid model, 3D recurrent convolutional neural networks (3DRCNN), for recognizing American Sign Language (ASL) gestures and temporally localizing their boundaries in continuous videos by integrating multi-modality features. The proposed model combined 3D convolutional neural network (3DCNN) for learning features from RGB, motion, and depth channels with an enhanced fully connected recurrent neural network (FC-RNN) capturing temporal information from short video clips, achieving a 69.2% accuracy on sequence videos for 27 ASL words in a newly collected ASL dataset, demonstrating its effectiveness in detecting ASL gestures from continuous videos [11].

Elboushaki et al. used MultiD-CNN as an approach for human gesture recognition in RGB-D videos, combining 3D ResNets and ConvLSTM to learn spatiotemporal features. The architecture simultaneously processes RGB and depth sequences, encoding temporal information into a motion representation, and employs a two-stream architecture for deep feature extraction. The study explores fusion strategies, showing that integrating multiple encoding methods enhances spatiotemporal feature learning with improved generalization capability [12]. Gokce et al. tackled the problem of Sign Language Recognition (SLR) by training separate 3D Convolutional Neural Networks (3D-CNN) for hands, face, and upper body regions, achieving improved accuracy through score-level fusion, with potential applications in Sign Language Translation (SLT) [13].

Transformers are developed to solve the sequence translation problem, that is any problem with input and output as sequences, including speech recognition problems, text-to-speech conversion, etc. These networks overcome the problem of memorizing long sequences in LSTM networks with the help of the concept of attention and hence they are very popular. This extraordinary ability of transformers has also attracted the attention of researchers in the field of sign language recognition and many researchers in this field use these networks. In an effort to continuous sign language recognition at the sentence level, Zhou et al. used the pre-trained transformer-based BERT network and the ResNet convolutional neural network with a full person video as well as hand frames for increasing the accuracy, which resulted in a good performance on different datasets [2].

Du et al. only used transformer-based networks to tackle the problem of sign language recognition [14]. In this work, to extract the spatial features of the images, a type of image transformer network (Swin transformer) was used, which extracts the features of the image with the help of the concept of attention. After extracting these features for all the frames of the sign language video, these features are transferred to another transformer to extract the temporal features and interactions of these features. Finally, using the Cross-entropy cost function, the loss was calculated. For the WLASL1000 dataset, they achieved an accuracy of 57.13%.

Alongside the global efforts dedicated to sign languages, a similar trend is observed in the advancement of Iranian Sign Language (ISL). Notably, ISL is a sophisticated language wherein both hand gestures and facial expressions play significant roles in conveying meaning within words and sentences. Ghanbari Azar et al. addressed the challenge of recognizing dynamic Iranian sign language words by employing the Hidden Markov Model [15]. Initially, they tracked the hand trajectory during gesture execution and extracted its features using spline interpolation. This system achieved 98% accuracy in recognizing 15 sign language words. Madani and Nahavi developed a system to recognize 20 sign language words, focusing on identifying isolated dynamic signs in Iranian sign language [16]. They employed adaptive mean shift to track the signer's hand, followed by feature extraction using the Radon transform and discrete cosine transform on the detected hand trajectory. Subsequently, four different classifiers, including the minimum distance classifiers, k-nearest neighbor algorithm, neural network, and support vector machine, were utilized for input class detection. Notably, the minimum distance method exhibited the highest accuracy of 95.56% among the tested subjects.

Rastgoo et al. proposed a hybrid model based on deep learning to recognize isolated dynamic Iranian Sign Language using video input of sign language performances [17]. This model comprises two main components: hand recognition and gesture recognition. The workflow begins with the detection of hands from the frames of the input video using Single-Shot Detector (SSD). Subsequently, three types of distinct features are extracted from the hand frames, including spatiotemporal features, hand joint positions, and shape and distance features of the hands. To extract spatiotemporal features, the pre-trained ResNet50 network was employed, while the model described in [18] was utilized to extract three-dimensional hand joint positions. The features concerning the shape and distance of the hands encompass components such as inclination and orientation. This model underwent training and testing on the isoGD dataset [19] and a set comprising 100 words in Iranian Sign Language, achieving an accuracy of 86.32% for the isoGD data. To address real-time application challenges, a straightforward and efficient model based on singular value decomposition of hand joint coordinates matrices was proposed. This model attained an accuracy of 99.5% on the Iranian RKS-PERSIANSIGN dataset and 86.1% on the isoGD dataset [20].

In this study, isolated Iranian Sign Language recognition was investigated using hand key point coordinates, elbow and wrist coordinates for each hand, lip key point coordinates, and the distance and angle between hands for each frame. These features were utilized in two separate networks with a late and early fusion of features. The exclusion of raw videos from the training process contributed to achieving high accuracies with limited data. Additionally, word embeddings were incorporated alongside the true class of the input to enhance training and improve the network's comprehension of the problem. Given the significant data requirement for training neural networks and the absence of an open-source dataset for Iranian sign language, collecting training data emerged as an important issue. To address this, a collection of word-level data, comprising 101 videos of words from individuals proficient in Iranian Sign Language, was gathered by our collaborators [21], which we will use in this research. Finally, the developed model was implemented in an interactive sign language training software to assess user performance. In summary, the main contributions of this research include:

- Extracting useful features from the body, hands, and face of the signer
- Introducing a hybrid word recognition network using extracted features
- Using the developed word-level model for sentence recognition
- Development of the first interactive Iranian Sign Language learning software with feedback to the user

## 2. Materials and tools

In this section, first, the dataset which is used in this study is introduced. This dataset was exclusively collected by our collaborators, as there was no open-source dataset for Iranian Sign Language. Also, given that training the network solely with raw image data necessitates a substantial amount of data, first, important features in sign language, such as the coordinates of the hands and lips key points, have been extracted and used for training the network. The details of this preprocessing procedure are thoroughly discussed in this section. Finally, the networks and methods used for training are explained.

*2.1. Dataset*

The data set used in this research contains videos in which only one single word is performed by a person fluent in Iranian Sign Language. All of this data was collected in the Islamic Azad University, Fereshtagan branch, which is a university specifically for students with special needs. Different backgrounds were used for each of the data, and a fixed background such as a green screen was not used. The reason for this is to make the data set closer to reality and more suitable for future work in investigating the use of the network in the real-world scenarios. This data set was recorded with a resolution of 600×800 and a frame rate of 25 frames per second. The words were chosen in such a way that they are most commonly used in academic environments such as universities. The number of performers in this dataset is 11, and the total data collected is 4040 videos. The average length of each data in this dataset is 57.01 frames, with a minimum length of 21 frames, and a maximum length being 116 frames. Additionally, with the start of the video of each word, the word is immediately initiated, and the end of each video of each word corresponds to the end of its performance.

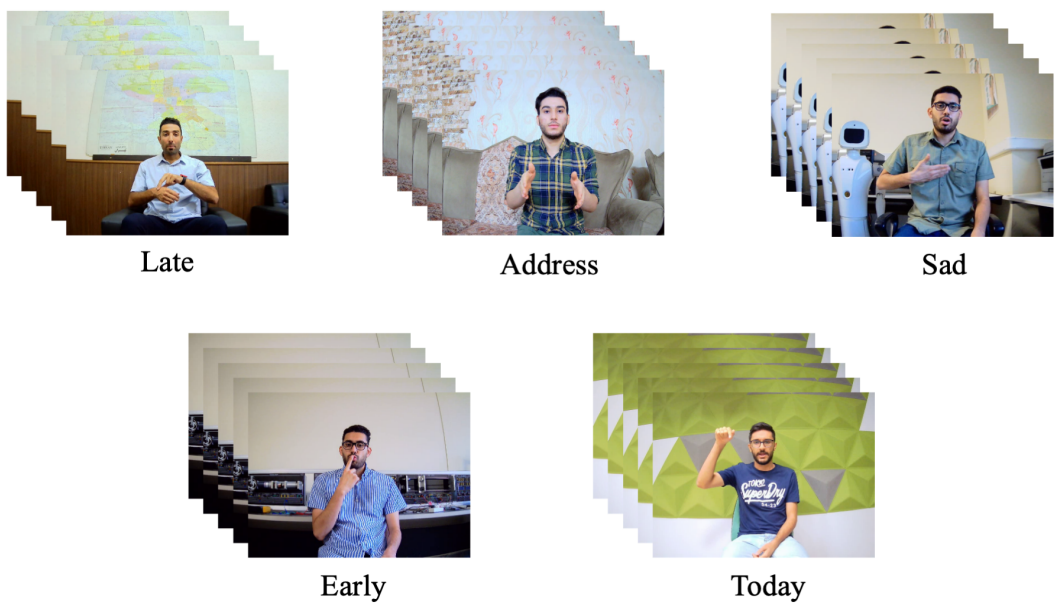

**Figure 1.** An overview of the word dataset.

To train the networks by this dataset, the dataset is divided into three parts: train, validation, and test. The data from 9 people was used in training, 1 person in validation, and 1 person in the test, and the selection of these people was done randomly. Figure 1 shows an overview of the dataset.

*2.2. Feature Extraction and Preprocessing*

The emergence of deep neural networks has reduced the need for pre-processing due to their high capability in feature extraction. Nevertheless, clean input data and meaningful features still offer many advantages including reduced training time, increased network accuracy, and the avoidance of complex networks. The pre-processing performed on our data involves the extraction of key and important features for sign language recognition. These features comprise the local coordinates of the fingers, the spatial coordinates of the key points of the hand, the coordinates of the lips, and the length and angle of the line connecting the two hands, all of which will be discussed in detail in the following sections.

*2.2.1. Input Length Correction*
An important point in training deep networks is the requirement to have consistent input length. As discussed earlier, the length of the videos in the sign language dataset varies from 21 to 116 frames. To address this issue, the length of 40 frames was chosen, slightly below the dataset's average frame count. This length was selected to be proximate to the average for word videos, ensuring it doesn't excessively burden computational resources nor compromise network accuracy. Then, all videos were adjusted to 40 frames. To achieve this, frames were randomly deleted if the video exceeded 40 frames, and zero-padded if the length of the video was less than 40 frames. Additionally, through the use of input masking, these zero-padded inputs will not influence network computations or results. It should be noted that randomly removing a number of frames for some data will help the network to be more robust and generalizable.

*2.2.2. Hand and Face Detection*
The hands and face of the person performing sign language are two very important components in recognizing sign language. Identifying these parts in the frames of the sign language video is helpful in this regard. Among the applications of hand and face recognition, it can be mentioned that it facilitates the extraction of features such as the key points of the lips and hands.

For this purpose, models are available that have the ability to recognize hands and faces. An important point is that there is no model for simultaneous recognition of hands and faces, and to use available models, we were forced to use two separate models, one for recognizing hands and another for recognizing faces. Using two separate models will increase the inference and feature extraction time. For this purpose, in this step, a single model will be trained for the simultaneous recognition of hands and face. This work, in addition to eliminating the need for us to use two models at the same time, allows us to get the desired result by labeling the training data by ourselves.

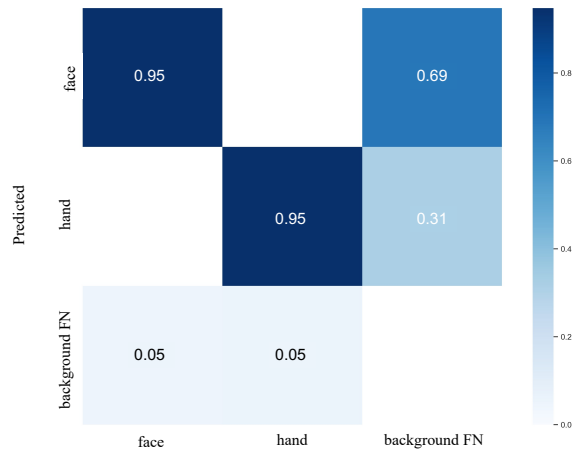

**Figure 2.** Confusion matrix of the trained YOLO model for test data.

The model used for training is YoloV5m, selected for its high accuracy and speed [22]. In this model, objects are detected using rectangular bounding boxes. Consequently, the label for each training data includes a vector comprising the values of the recognized object class, the normalized coordinates of the center of the bounding box, the length, and the width of the bounding box, which was prepared manually for each of the data that was selected. To train this network, a combination of 4 different datasets was used. These datasets include roboflow hand data [23], roboflow-FAST-model face data [24], several frames from Iranian deaf news, and finally some frames from our Iranian Sign Language word data. The dataset comprises 8543 images, divided into test, validation, and training datasets at an 8:1:1 ratio.

This model was trained with a batch size of 16 and for 150 IPACs, and finally, the best weights according to the cost function of the validation data were selected as the weights of the main model. The confusion matrix for the test data is shown in Figure 2, and an example of the model's detection of unseen data is presented in Figure 3. This model has the ability to recognize hands and faces in photos and videos, which will help us in feature extraction.

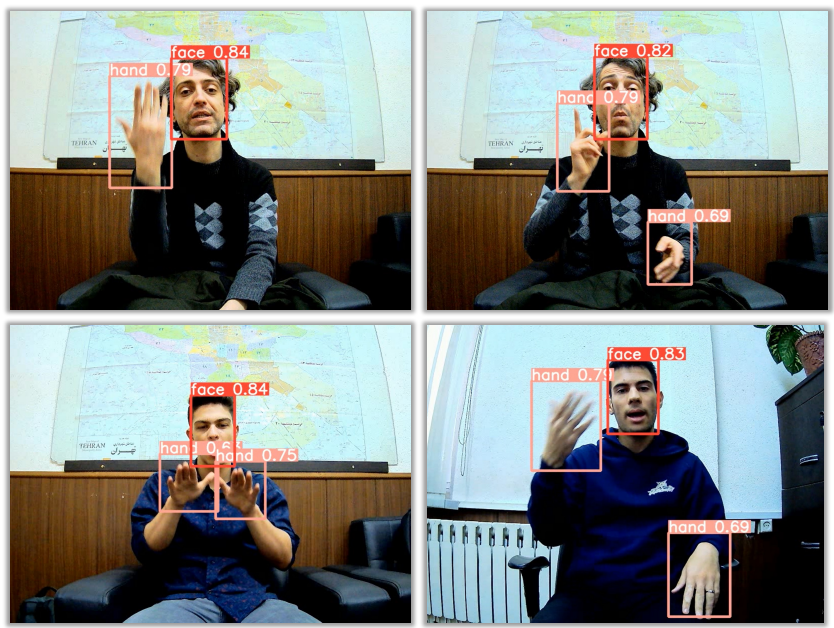

**Figure 3.** Recognition results for some instances using the trained YOLO model.

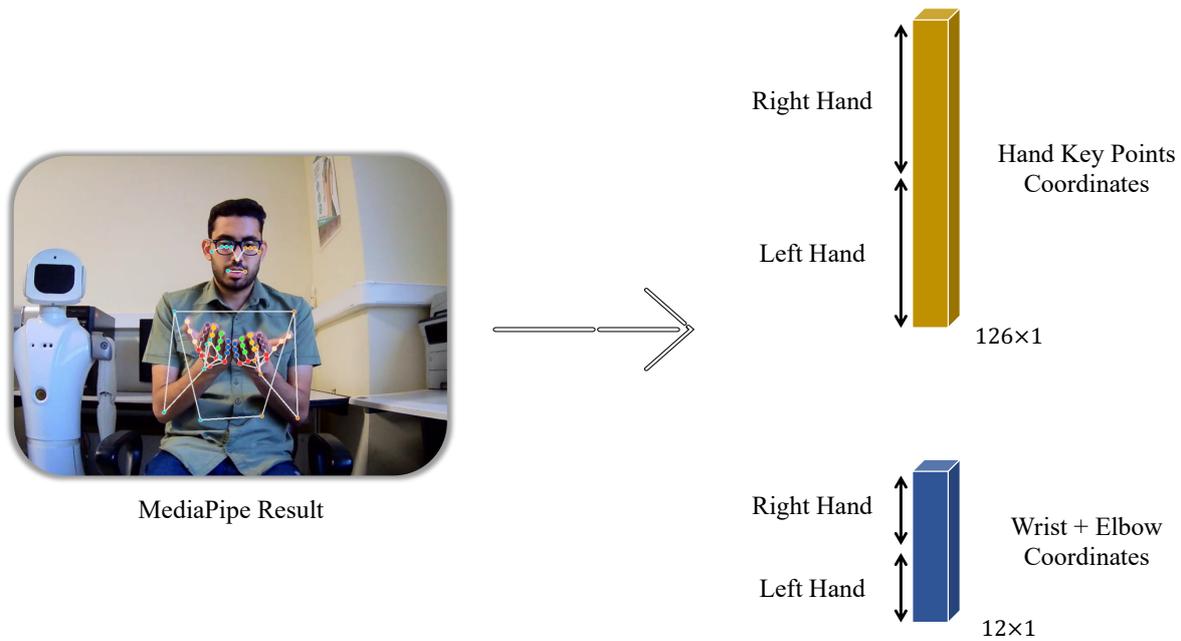

**Figure 4.** Feature extraction of hand key points and wrist and elbow coordinates

*2.2.3. Hand Key points Extraction*

Hands play the most important role in sign language. It is almost impossible to recognize sign language words without access to hand gestures. For this reason, extracting rich features from the user's hands is very important. In this research, MediaPipe was used to extract the coordinates of the key points of the hands [25]. MediaPipe has the ability to extract body key points from images and videos. For example, we can extract important joint points such as the head, hands, feet, and other parts of the body in three dimensions.

In the first step, MediaPipe extracts hand key points from all frames of each sign language video. These three-dimensional coordinates are extracted locally for each hand, where the origin of the moving coordinates is located at the geometric center of the hand, moving along with the hand's movement. This is to ensure that this information only includes the shape of the hand and removes the hand's movement in space as a factor. The number of these points for each hand is 21, resulting in a total of 126 features, covering two hands and three components (x, y, and z). This feature vector will be utilized for network training along with other extracted features.

Since in the previous step, only hand shape information was extracted, and not the general movements of the hand, these movements are also extracted in this step. For this purpose, the moving point of two points of each hand (elbow and wrist) which, like before, includes three-dimensional coordinates and with the origin of the center of the person's pelvis was extracted for each frame with the help of MediaPipe. This feature vector was also extracted for two hands for a total of 4 four points and 12 features. Figure 4 shows a summary of the extracted features discussed.

*2.2.4. Lips Key Points Extraction*

Since lip-reading and lip movements are very important in sign language, we will use this feature to recognize sign language in our system. For this purpose, the MediaPipe tool has been used again, the output of which will be 40 points for the lips (including 3D coordinates for each point), which will eventually result in a vector with the size of 120 features. If MediaPipe alone is used to extract these points from the frames of the dataset videos, a good performance will not be achieved, and sometimes the points for lips are not detected. To solve this problem, first, using the hand and face detection model that was mentioned earlier, the cropped image of the signer's face was obtained from each frame, and then this image was used in MediaPipe to extract the points of the lips. With this operation, the previous problem was completely solved, and points were obtained well for all frames (see Figure 5).

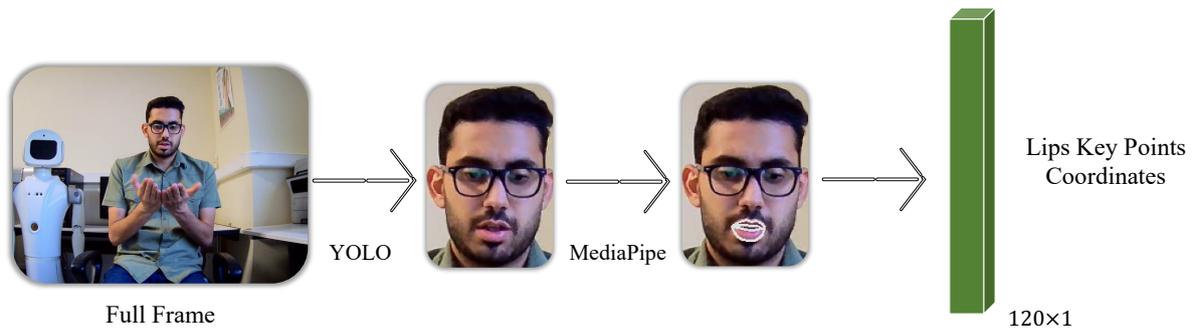

**Figure 5.** Feature extraction of lips key points

### 2.2.5. Relative Position of Hands Features

The position of the hands relative to each other is also one of the factors that can make a difference between the executed signs. For this reason, in this part, two features of the distance between two hands and their angle were used.

For each hand, our hand and face recognition model will predict a rectangle within which the hand will completely fall inside. The distance between the centers of these rectangles for each hand was chosen as a measure of the distance between the hands. In this way, first, the coordinates of the center of each rectangle for each hand were normalized with the length and width of the image, and then the Euclidean distance of these two normalized points was calculated. It should be noted that if a hand is not present in the image, the center of the hypothetical rectangle selected for that hand was chosen to be the center of the last recognized rectangle for that hand, and if there was no history of the detection of that hand, the lowest point of the center of the image was chosen as the center of this hypothetical rectangle.

In addition to the distance between the two hands, their angle with the horizontal line was also obtained with the help of the centers of the rectangles surrounding the hands. The angle, together with the distance between the two hands, which form a two-dimensional feature vector for each frame, was used to train the model along with other features. We see an example of this feature in Figure 6.

It should be mentioned that for all the aforementioned features, if the model was not able to recognize the desired feature or if that feature did not exist, then the zero vector was returned as the output of the feature extractor.

### 2.3. Networks and Training

In this section, we will delve into the details of the trained model and training parameters. Since the input data are time series, a sensible choice would be to employ RNN modules (such as LSTM or GRU) or, alternatively, newer modules such as transformers. Due to the greater memory capacity of the transformer compared to RNN networks, as well as their faster processing speed and the possibility of training transformers with more parameters (due to the parallel processing in transformers), we opt for these networks as the primary foundation for sequence modeling.

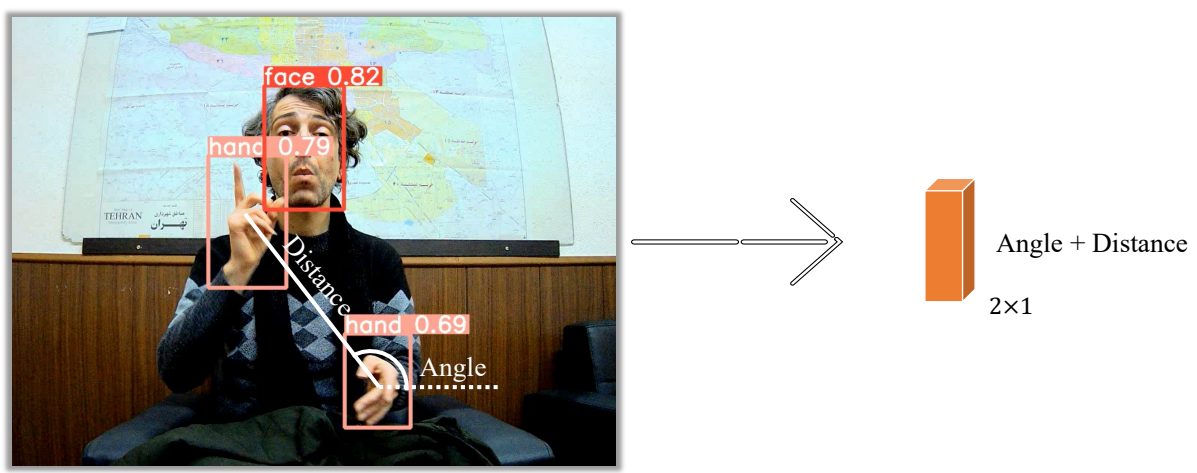

**Figure 6.** Feature extraction of angle and distance between hands

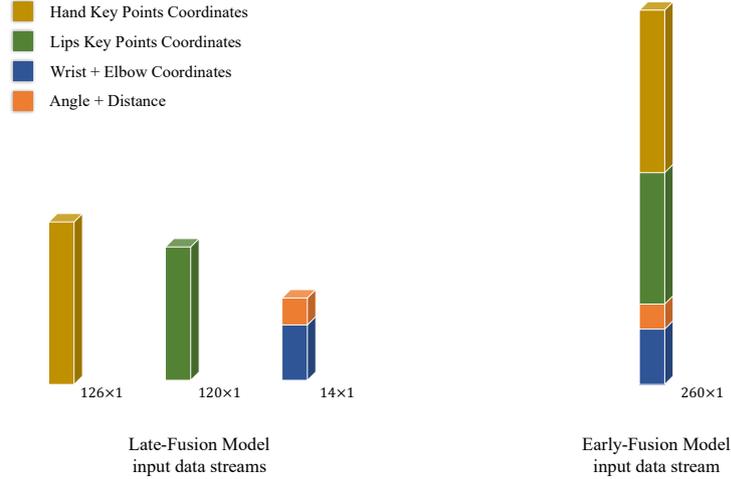

**Figure 7.** Data structure of our input features to the networks.

*2.3.1. Data Structure*

To train our model, we will utilize the features extracted in the previous sections. These features include coordinates of hand key points, elbows, and wrists for each hand, lip key point coordinates, and the distance and angle between hands for each frame. These features form three different input streams for the first network, with the data being merged at the end of the network (the late fusion network). These data streams comprise hand key point vectors, lip key point coordinates vectors, and combined elbow and wrist coordinates vectors with the distance and angle of the hands. For the second network, where data is integrated from the beginning, all features are combined into a single vector that enters the model (early fusion model). Refer to Figure 7 for the structure of the data.

*2.3.2. Late Fusion Model*

Late fusion models refer to models that merge the input data flows together toward the end of the network. Each data stream can have unique information and characteristics, and transferring this information to the model can significantly improve the performance of the model. In addition, late fusion models are able to combine the best aspects of each data stream to improve model accuracy and performance. For example, in image processing, a late fusion model can use different streams of images, such as the RGB and depth images, and extract and combine different features from them. This can help the model learn more details in the image and improve the accuracy and quality of the prediction.

In this model, three transformer encoder modules were used for each data stream, and finally, one transformer encoder module was used for the integrated data. The three data streams introduced earlier enter the transformer modules separately, and the output of each module comprises vectors with a size equal to the input vectors. These vectors contain information from the frames preceding and succeeding them. In effect, these primitive encoders examine the relationship of each data stream to itself and output richer vectors.

After the context-aware vectors were formed by the first encoders for each data stream, all of these data were merged together to form a single vector per frame through concatenation. Now, for the last step, these comprehensive vectors entered the last layer of the transformer encoder, and the output of this encoder entered the dense layer with the number of neurons equal to the number of word classes. This output will be compared with the one-hot data vectors, and then the model will be updated (Figure 8).

In addition to training the network with true word labels, represented as one-hot vectors, this model was also trained with pre-trained FastText embeddings on the Persian Wikipedia data corresponding to each word. These embedding vectors for each word were obtained using the skip-gram architecture, with dimensions set to 300 for each word. Incorporating word embedding as an auxiliary task provides additional information about the semantic relationships and contextual similarities between sign language words to the deep network. This additional knowledge helps the network better understand the nuances and complex patterns inherent in sign language movements. By using the rich semantic representation encoded in word embeddings, the network achieves a deeper understanding of the meaning and lexical structure of sign language, thereby enhancing its ability to distinguish between similar signs that belong to different word categories. Furthermore, multi-task learning enables the network to use joint representations between the primary task of one-hot vector classification and the secondary task of word embedding prediction. This mutual advantage arises from the fact that both tasks share the same network architecture and hidden layers. As a result, the network can take advantage of synergies between tasks, effectively regularize the learning process, and reduce the risk of overfitting.

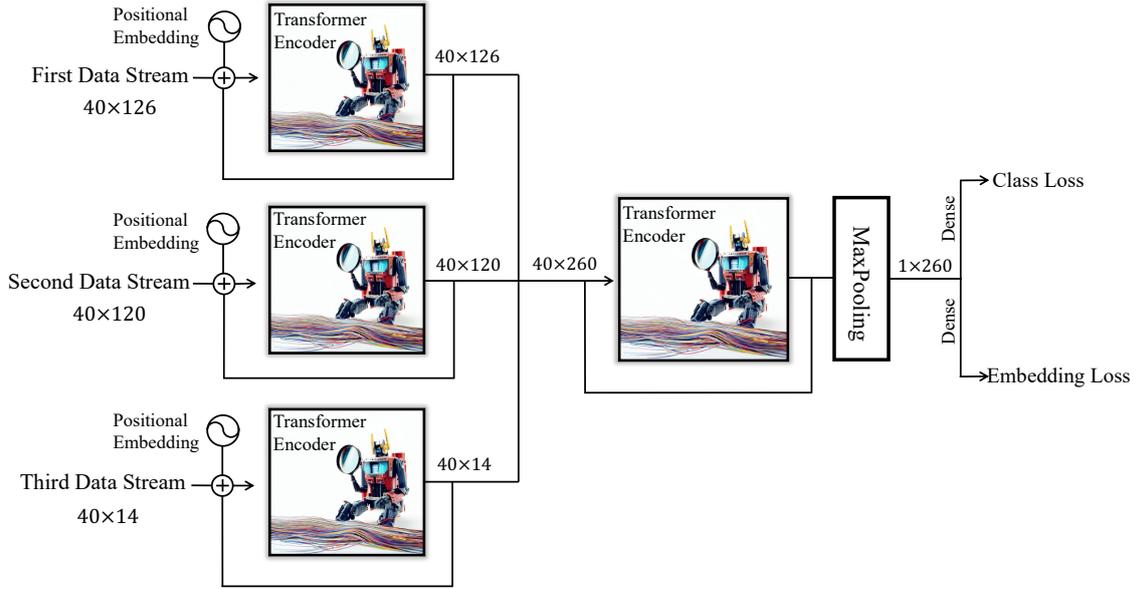

**Figure 8.** Our late fusion model structure. Each transformer encoder has the same structure as the original one introduced in [26].

In the late fusion model, each of the 4 encoders had 12 heads, and the number of dense layer neurons in every encoder was also selected for the hand key points encoder, elbow and wrist coordinates encoder, and lip key points encoder, respectively 256, 256, and 64 neurons. The number of neurons for the final encoder handling integrated vectors was set at 512 neurons. Moreover, all encoders featured a skip connection from their input layers, which facilitated the transfer of information and gradients during training. This model had a total of 5,183,851 trainable parameters, and the ratio of data to parameters was equal to $6 \times 10^{-4}$.

Softmax activation function was used for the word label class, and linear activation function was used for the embedding output. This model utilized two cost functions: Categorical Crossentropy for word label class, and CosineSimilarity for embedding training. The final cost function comprised the weighted sum of these two functions, with coefficients of 1.8 for the word label class cost function and 0.5 for the embedding cost function.

The implemented optimizer was chosen to be Adamax with a learning rate of 0.0012 and a weight decay rate of 0.0001. Top-1 and Top-5 accuracies were also employed as metrics. The model was trained for 200 epochs and after the completion of the training, the weights corresponding to the epoch that had the highest Top-1 accuracy on the validation data were selected as the final weights of the model. Additionally, a rate of 0.15 for label smoothing was considered.

*2.3.3. Early Fusion Model*

In addition to the previous model where the data streams were merged at the end of the network, another model was presented in which the data are merged from the beginning and enters the network in the form of a single vector for each frame (Figure 9). In this model, only one transformer encoder layer was used, and as before, there are two different cost functions for class training and word embedding. The reason for training this model is to use it alongside the previous model in ensemble learning to acquire higher accuracy. These two models predict the target word class in different ways and are suitable for ensemble learning. More details about the final model, which is a combination of these two models, will be presented in the following sections.

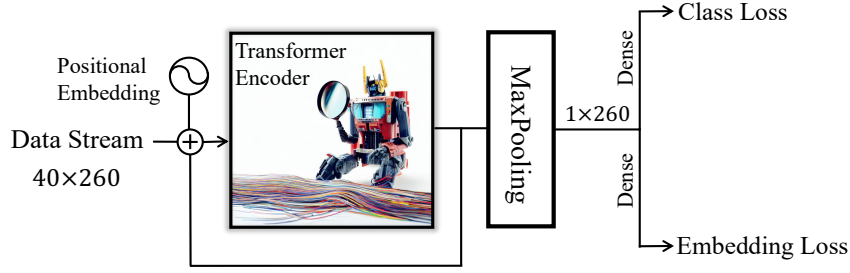

**Figure 9.** Our early fusion model structure.

The encoder used in this model had 12 heads, and the number of neurons in its dense layer was 512. Similar to the previous encoder, there was a skip connection from the input. This model had a total of 3,584,791 trainable parameters, with the ratio of data to parameters equal to $9.48 \times 10^{-4}$. All other settings, including the optimizer, cost functions, and training settings, were chosen exactly identically to the late fusion model.

*2.3.4. Final Model*

Ensemble models have different types. Some of them do not have learning abilities, such as voting or averaging, while others do. In this research, the latter method was used. More specifically, the weights of the early and late fusion models remained constant, but several dense layers were added after the models, and these layers were retrained with our data.

In this model, the class outputs of the two previous models were concatenated and formed into a single vector. Then this vector was passed through multiple dense layers, and finally, in the last layer, the dense layer had the same number of neurons as the number of classes. Finally, this model was trained on the dataset again. The number of layers and neurons in each layer was also obtained with the help of the genetic algorithm.

The optimizer used is Adamax, with a learning rate of 0.0015 and a weight decay rate of 0.0004. Top-1 and Top-5 accuracies are also used as metrics. The model was trained for 100 epochs, and after training, the weights corresponding to the epoch that had the highest Top-1 accuracy on the validation data were selected as the final weights of the model.

*2.3.5. Genetic optimization of the network*

For determining the optimal structure of the added layers in the final model, the number of added layers and the number of neurons in each layer were obtained using genetic algorithm. Since the accuracies of different models are close to each other, to better distinguish between the values of the objective function, it is defined as follows:

$$f = e^{\frac{acc_{validation}}{2.5}}$$

It should be noted that, in order to prevent the model from becoming too large, a maximum of 8 layers and a maximum number of 756 neurons are considered in the calculations.

The chromosomes used in this research had 9 genes. The first gene specifies the number of layers and can have a value between 1 and 8. The following genes all represent the number of neurons in each layer, which can have values from 1 to 756. It should be noted that the values of the genes corresponding to the number of neurons in each layer are zero in the absence of that layer.

The process of selecting parents in this algorithm was done randomly. The probability of selecting each chromosome was calculated based on the normalized value of the objective function compared to the sum of the objective functions. In this method, chromosomes that rank higher than others are more likely to be selected as parents. However, the possibility of choosing parents with lower costs is not excluded, so there is higher chance to reach a global solution. In each step, 10 parents are selected in this way and offspring are produced from the intersection of the chromosomes of these 10 parents.

To produce a new generation, uniform crossover was used. In uniform crossover, each gene is randomly selected from one of the corresponding genes of the parent chromosomes. One thing that should be noted is that in chromosomes, the values of some genes corresponding to a layer that does not exist will be zero, and this should be considered in the crossover of genes. To solve this challenge, the process will be done in this way: during the crossover of chromosomes, as long as both parent chromosomes have non-zero genes, for each child gene, we randomly use the corresponding genes of the parent, and for genes from parents that have only one non-zero gene, its value will be exactly copied to the child. In this way, valid children are produced.

Additionally, the mutation rate for the number of layers gene is 0.5% for each gene, and for the mutation in the number of neurons, the value is 0.1% for each gene. It should be noted that if the number of layers jumps to a number higher than its current value, the number of new neurons (which were zero before) will be randomly selected. Also, in this method, the best chromosome does not undergo the mutation process, and the best solution will always be preserved in each generation of mutation.

Also, randomly and with a probability of 8%, a completely random chromosome will replace the chromosome of the previous generation that has the lowest objective value. The number of generations should be determined so that considering the time-consuming calculations, the required time is reasonable and we also get close to the optimal solution. Therefore, in order to be able to finish the algorithm in a reasonable time, the number of generations was chosen to be 30, and the stopping criterion for the algorithm was reaching the 30$^{th}$ generation.

*2.4. Software Development and Assessing Acceptability*

In today's society, effective communication is essential. Sign language serves as a vital medium of communication between the deaf community and facilitates interaction and understanding. However, mastering sign language requires dedicated practice and guidance. This is where easy-to-use sign language training software becomes essential. In this part, sign language training software with feedback to the user through the evaluation of the signs implemented by the developed models is introduced. This is the first step towards automated sign language evaluation in learning software, potentially having a significant impact on learners.

Such sign language training software also partly solves the problem of a lack of qualified sign language instructors. Finding skilled sign language instructors can be challenging, especially in areas with limited resources or a small deaf community. By using software that can evaluate user performance, people interested in learning sign language can access quality education regardless of their geographic location. This increases the accessibility and availability of sign language education and empowers more people to learn and communicate.

After developing the software and integrating the trained models, we conducted a user study to evaluate the impact of our model within the software. A total of 30 individuals participated in a software demonstration followed by a survey. This group included 15 people without hearing impairments (Group 1) and 15 people who are deaf or hard of hearing (Group 2). Upon reviewing the research and questionnaire conducted in [27], we concluded to utilize a Persian translation of the UTAUT questionnaire, incorporating 9 out of 13 items, to assess the acceptability and usefulness of our work. Additionally, two extra items were included in the questionnaire to ensure comprehensive coverage of all study aspects. The questions for each item were crafted in line with the UTAUT framework, with minor adjustments tailored specifically for this study's context (Table 1).

Except for the user feedback question, the participants should rate the items on a five-point Likert scale (ranging from 1 to 5). The scale included verbal anchors ranging from "very low/totally disagree: 1" to "very much/totally agree: 5", allowing the subjects to express their opinions on the questions/items.

**3. Results**

*3.1. Network Performance*

*3.1.1. Single word evaluation*

The optimization process of the layers and the number of neurons in the layers that were added to the ensemble model reached the top chromosome after 30 generations. The final network consisted of 6 dense layers, with the number of neurons in the layers being 310, 693, 465, 638, 513, and 406, respectively.

First, the accuracy obtained on the late and early fusion models alone was examined. For the late fusion model, the Top-1 accuracy was 88.8% and the accuracy of the Top-5 predictions was 98.8%. Also, for the early fusion model, these accuracies were 85.4% and 94.18% for Top-1 and Top-5 respectively. As expected, the late fusion model had a better performance, but the difference in the procedure to reach the final prediction in networks may increase the overall performance of the final ensemble model that leverages both models' capabilities. Meanwhile, the average prediction time of each word by this network was 47 milliseconds for early fusion and 56 milliseconds for the late fusion model.

The optimized ensemble model had 2,311,918 trainable parameters and after being trained on the available training data, the final model was obtained. The Top-1 accuracy of the final model on the test data was 90.2% and the accuracy of its Top-5 predictions was 93.1%, which is comparable to some recently developed networks for similar datasets. The use of the combined model brought us to an accuracy that we could not achieve with any of the single models, which is very important and shows the power of ensemble learning. It should be noted that the average prediction time of each word by the network was equal to 100 milliseconds, which makes it suitable for real-time use.

**Table 1.** Our proposed UTAUT-based questionnaire

| | | | |
|---|---|---|---|
| Anxiety (ANX) | If I should use the software, I would be afraid to make mistakes with it. | If I should use the software, I would be afraid to break something. | |
| Attitude Towards Technology (ATT) | I think it is a good idea using this software. | I don't think that this software can be as effective as humans. | |
| Facilitating Conditions (FC) | I have everything I need to use this software. | I know enough of the software to make use of it. | |
| Intention To Use (ITU) | I think I will use the software during the next few days. | | |
| Perceived Adaptiveness (PAD) | I think the software can be adaptive to what I need. | | |
| Perceived Enjoyment (PENJ) | I enjoy using the software. | I find the software enjoyable. | I find the software boring. |
| Perceived Ease of Use (PEOU) | I think I will know quickly how to use the software. | I find the software easy to use. | I think I can use the software without any help. |
| Perceived Usefulness (PU) | I think the software is useful to me. | I think the software can help me with learning. | |
| Trust | I would trust the software evaluations. | | |
| AI Effectiveness | I believe the AI features in the software can enhance my sign language learning efficiency. | | |
| Instructor Need | This software can minimize the necessity for an instructor. | | |

The effect of adding each feature can also be seen in Table 2. As you see, the most discriminative feature is hand key points representing the shape of the hands, but by using this feature alone, accuracies higher than 76.7% are not reachable. Another notable point is that although lip coordinates perform very poorly when using them alone, when used with hand key points, they increase the accuracy a lot. Finally, the performance of LSTM-based networks is shown, where the exact same network architecture as before was used except all transformer modules were replaced by LSTM ones. This clearly shows the advantage of transformer networks over LSTM in our case.

*3.1.2. Continuous evaluation*

In the preceding section, the model's performance was assessed using individual words from our dataset. This scenario provided the model with clean, well-defined data, where each word began precisely at the start of the video and concluded at its end. However, such ideal conditions are rarely encountered in real-world scenarios, particularly in continuous sign language recognition, where multiple words may exist within a single segment and their boundaries are not clearly specified. Despite not being trained explicitly for such scenarios, this section evaluates the model's performance under similar conditions.

To achieve this, we constructed sign language sentences by concatenating words from our dataset. Each resulting sentence comprised a sequence of words with the end of one word seamlessly attached to the beginning of the next. While this method introduces interruptions and discontinuities at connection points, it also ensures that sentences generated are similar to those formed from individual words, facilitating a fair assessment of the model's capabilities. Moreover, the recognition model, initially designed for single words, holds promise for applications in sign language teaching software. In our work, we developed an automated word evaluator for the software, which could also assess entire sentences and provide feedback on incorrectly performed words for future works—a functionality lacking in models directly predicting sentences.

Table 2. Effect of different features inclusions and transformer modules on Top-1 and Top-5 accuracies (Top-5 in parentheses, reported numbers are percentage.)

| Ablation Settings | | | | Performance | | |
|---|---|---|---|---|---|---|
| Hand Key points | Lips Key points | Wrist and Elbow + Distance and angle | Temporal Module | Early Fusion | Late Fusion | Ensemble Model |
| ✓ | ✗ | ✗ | Transformer | 76.7 (90.3) | - | - |
| ✗ | ✓ | ✗ | Transformer | 14 (28) | - | - |
| ✗ | ✗ | ✓ | Transformer | 35.3 (65.4) | - | - |
| ✓ | ✗ | ✓ | Transformer | 81.9 (**94.1**) | 82.9 (96.8) | 85.2 (91.2) |
| ✓ | ✓ | ✗ | Transformer | 82.3 (93.4) | 84.1 (97.3) | 84.8 (**93.6**) |
| ✓ | ✓ | ✓ | Transformer | **85.4** (**94.1**) | **88.8** (**98.8**) | **90.2** (93.1) |
| ✓ | ✓ | ✓ | LSTM (1 layer) | 61.2 (88.2) | 68.9 (89.1) | 68.2 (78.2) |
| ✓ | ✓ | ✓ | LSTM (2 layers) | 70.9 (89) | 66.9 (88.5) | 73.9 (81.5) |

Twenty sentences were prepared using words from the test dataset. A moving window with a size of 40 frames was applied to the data and fed into the network, yielding a one-hot vector indicating the probability of each word's occurrence for each window. Words with a probability exceeding a predetermined threshold (set at 0.2 in this study) were considered valid recognitions, and the others were labeled as "null". Upon processing, the window advanced by a step length of 5 frames. If the recognized word matched the previous one, or if it was similar but separated by one or more "null" values, it was disregarded. Only confidently recognized new words, distinct from the previous one, were accepted, which is a methodology similar to prior research efforts [28]. The performance of the model on these constructed sentences is summarized in Table 3.

Table 3. Results of the continuous sign language recognition scenario. (I: Insertion, D: Deletion, S: Substitution)

| Sentence index (# of words) | Mean confidence | Number of errors (I, D, S) | Sentence index (# of words) | Mean confidence | Number of errors (I, D, S) |
|---|---|---|---|---|---|
| 1 (3) | 0.53 | 1 | 11 (3) | 0.72 | 0 |
| 2 (2) | 0.41 | 1 | 12 (4) | 0.43 | 2 |
| 3 (3) | 0.72 | 0 | 13 (2) | 0.81 | 0 |
| 4 (3) | 0.6 | 0 | 14 (4) | 0.37 | 1 |
| 5 (4) | 0.62 | 1 | 15 (4) | 0.31 | 1 |
| 6 (5) | 0.39 | 2 | 16 (4) | 0.59 | 2 |
| 7 (3) | 0.57 | 0 | 17 (4) | 0.57 | 0 |
| 8 (4) | 0.51 | 0 | 18 (3) | 0.42 | 1 |
| 9 (4) | 0.61 | 1 | 19 (3) | 0.55 | 0 |
| 10 (3) | 0.67 | 0 | 20 (4) | 0.63 | 0 |
| Average (total) | 0.551 | 0.65 | | | |

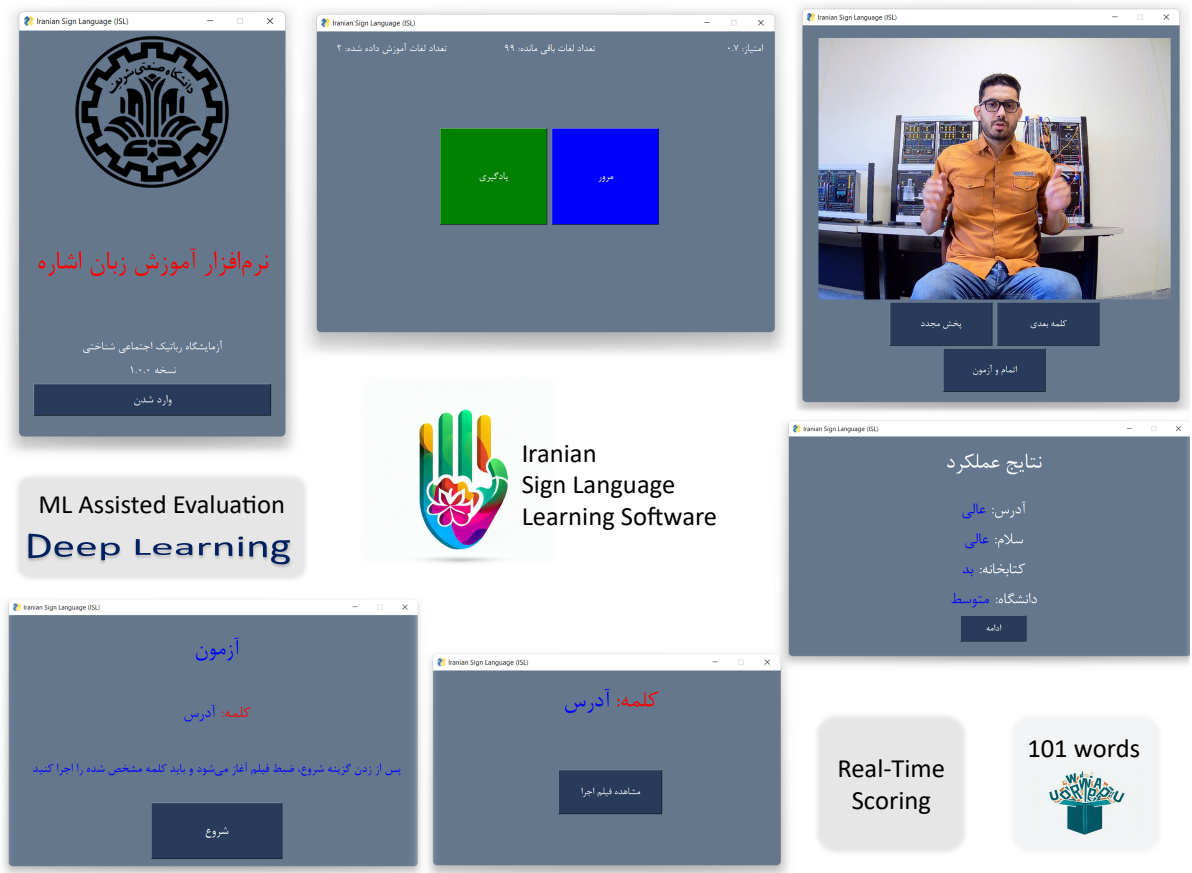

**Figure 10.** An overview of the Iranian Sign Language learning software.

## 3.2. Real World Implementation

### 3.2.1. Software Details

To develop the sign language teaching software (Figure 10), the PysimpleGUI Python library was used, which provides the ability to create a graphical user interface. After launching the software, the initial page containing the logo and title is displayed to the user. They can click the "Login" button to go to the next page. On the second page, an overview of the learning progress is presented to the user. It displays information such as the number of words learned, the remaining words, and their score. This score is assigned based on the user's performance in performing the signs. Next, the user can choose between two options: "Learn" or "Review". The review option re-displays the previously learned words to the user with priority given to the user's performance on them.

In the learning section, a specific word will be displayed to the user. This word is displayed as text, and the user can view the visual representation of the sign (which is one of the items in the word dataset) by clicking on the "Watch Video" button. In this section, after showing the word performance video, the user can do one of these three things: 1- repeat the video, 2- go to the next word to see the video of it, and 3- complete the learning process and take a test on the previous words.

If the "Finish and test" option is selected, a word is presented to the user, and they are asked to execute it in sign language. By clicking the "Start" button, the software will start recording video using the computer's camera. The user executes the word while viewing the camera output in real-time. After completing the word execution, the user can click on the "Finish" button to end the recording. Then the same process is repeated for the next word. At the end, the user's performance for each word will be displayed. This feedback is derived from the accuracy of the model in recognizing the desired word. It should be noted that the time required to process each word executed by the user is 6.8 seconds on average, which is mainly caused by preprocessing.

Furthermore, precise timing between the user's initiation of signing by selecting the start button and concluding by selecting the finish button is not necessary. A moving window with a size of 40 frames will be applied to the recorded video, allowing flexibility in timing. The highest confidence prediction among the windows will then be selected as the predicted word.

Table 4. A summary of the user study results and statistical analysis of the questionnaire items. (Group 1: people without hearing impairments, Group 2: deaf or hard of hearing people)

| Category | Mean (std) | | T-statistic | P-value |
|---|---|---|---|---|
| | Group 1 | Group 2 | | |
| ANX | 1.97 (0.91) | 2.4 (1.05) | -1.51 | 0.14 |
| ATT | 4.6 (0.49) | 3.73 (1.29) | 2.3 | **0.037** |
| FC | 3.83 (0.97) | 3.5 (0.92) | 1.17 | 0.25 |
| ITU | 3 (1.15) | 3.26 (1.29) | -0.52 | 0.61 |
| PAD | 3.4 (1.02) | 3.13 (1.26) | 0.59 | 0.56 |
| PENJ | 3.6 (1.2) | 3.06 (1.14) | 2.45 | **0.018** |
| PEOU | 4.4 (0.88) | 3.93 (1.2) | 2.36 | **0.02** |
| PU | 4.23 (0.71) | 3.8 (0.83) | 2.12 | **0.035** |
| Trust | 4.13 (0.72) | 3.26 (1.06) | 2.82 | **0.013** |
| AI Effect | 4.53 (0.5) | 4 (0.51) | 2.25 | **0.04** |
| Need for Instructor | 3.66 (1.19) | 3.26 (1.38) | 1.24 | 0.23 |

*3.2.2. User Study Results*

In this part, we will go through the user study results, as shown in Table 4. Results indicate that Group 1 generally rates the software higher in every item (except for a slight difference in ITU), with almost half of these differences being statistically significant. Categories such as Attitude Towards Technology (ATT), Perceived Enjoyment (PENJ), Perceived Ease of Use (PEOU), Perceived Usefulness (PU), Trust, and AI Effect exhibit small p-values (<0.05) indicating meaningful differences between the two groups' responses.

Users without hearing impairments (Group 1) reported a significantly more positive attitude towards the software, finding it more enjoyable and easier to use. They also perceived the software as more useful and trustworthy, appreciating the AI integration to a greater extent. These users exhibited higher confidence in the software and were less anxious while using it. This suggests that the software is well-received by users without hearing impairments, who find it user-friendly, effective, and enjoyable.

One interesting finding is that Group 2 exhibits less trust in AI evaluations and perceives AI integration as less effective compared to Group 1. This could be attributed to Group 2's familiarity with Iranian Sign Language, understanding its complexities, and their doubts regarding AI's ability to comprehend it (which shows its effect in "need for instructions" item too.). Another potential reason for this might be that, based on some other research [29], individuals with disabilities may have valid reasons to be cautious or distrustful of AI systems, especially when fairness and privacy are not adequately addressed.

While Group 1 generally had higher scores, indicating a more positive reception, the fact that Group 2's scores are still relatively high suggests that the software is well-received among deaf or hard-of-hearing users too. This demonstrates the potential for further improvements to enhance user experience for this group.

We also collected user feedback for the software. Most of the feedback from Group 1 focused on the user interface, suggesting improvements to make it more visually appealing. In contrast, Group 2 provided more technical feedback. They suggested integrating the software with games to enhance engagement and user motivation. Additionally, there was a suggestion that the software, with its current capabilities, should target individuals without hearing impairments who are interested in learning Iranian Sign Language, as deaf or hard-of-hearing people generally know the sign language at this level. Lastly, it was noted that a single word in Iranian Sign Language can have multiple signs, a complexity not currently addressed by the software.

## 4. Discussion and Limitations

In our word recognition model, each input is assigned a word from the dataset. However, if a word outside this dataset is performed, the model incorrectly predicts one of the existing dataset words. To address this, adding an additional class for managing out-of-dataset words is desirable for future work. Additionally, in our dataset, words precisely start and end with the corresponding sign execution, which should be considered during usage. Although windowing partially mitigates this issue, further refinement is needed.

Moreover, for sentence recognition, we used data with discontinuities at word borders, and fully continuous sentences were not tested. This limitation should be explored in future studies to enhance sentence recognition capabilities. The fact that we are aiming to translate the sentences word by word will help us in the future to give feedback to users who are performing a sentence based on each word. Some other post-processing techniques like using LLMs might be used to convert translated sentences to natural language. Also, for future improvements, it is recommended to increase the dataset size and diversity. Although the number of existing classes is sufficient, expanding the vocabulary will enable more diverse sentence translations.

A notable aspect of our sign language recognition software is the average delay of 6.8 seconds per word recognition. This delay is due to the non-optimal software language and the heavy computational load of the models, particularly in the feature extraction part. Investigating lighter and more concise features could reduce computational load and improve performance.

Enhancing the user interface to be more professional and user-friendly is another critical step, as was mentioned frequently in the user study. Adding the ability to communicate with and be monitored by sign language teachers would also be beneficial. An essential feature for development is teaching sentences and providing feedback on user-executed sentences. For example, after teaching a set of words, users should be asked to construct a sentence using these words and then receive detailed feedback on their execution for both the entire sentence and individual words. Based on our user study feedback, it is also desirable to incorporate some games in the software and also consider words with multiple signs.

## 5. Conclusion

Sign language is the primary means of communication for millions of deaf individuals worldwide who use it continuously. For this reason, this language is of great importance and should receive special attention. In Iran, although there are no official statistics, it is estimated that there are about one million deaf and hard-of-hearing people. So far, efforts made for Iranian Sign Language have been very limited, and there is a strong need for more effort and development in this area. Additionally, there is a global need for more accurate video-based models for sign language translation, with the practical implementation standing out as a notable gap in current research. In this research, a word-level Iranian Sign Language recognition model was developed and trained on a dataset with 101 words. Features such as hand key points, lip coordinates, elbow and wrist positions, and hand orientation and distance were extracted using MediaPipe and YOLO models trained on sign language videos and used for training our model. Two late fusion and early fusion models, achieve accuracies of 88.8% and 85.4% on test word data, respectively. The ensemble model which is the combination of the two previous models and with optimized parameters through genetic algorithms, achieves a high accuracy of 90.2% on the test dataset. This model aims to support software development for sign language learning. Additionally, the model demonstrated promising results with an average 0.65 error rate when applied to 20 sentences composed of concatenated word videos.

Finally, our developed Iranian Sign Language software, which is the first of its kind, integrates this model to provide real-time feedback on users' sign execution. This interactive tool can teach 101 words and evaluate user performance using the model. Our user study indicated positive acceptance and utility, particularly among individuals without hearing impairments. However, it was observed that deaf and hard-of-hearing individuals exhibited lower trust in AI and its utility. This software represents an initial step toward the practical implementation of such technologies, offering increased opportunities, especially for people with disabilities.